\documentclass[journal]{IEEEtran}

\usepackage{times}
\usepackage{epsfig}
\usepackage{graphicx}
\usepackage{amsmath}
\usepackage{amssymb}
\usepackage{xcolor}

\usepackage{subfigure}
\usepackage{multirow}
\usepackage{array}

\begin{document}

\title{Deeply Activated Salient Region for Instance Search}

\author{Hui-Chu Xiao,
        Wan-Lei Zhao*,
        Jie Lin,
        and Chong-Wah Ngo,~\IEEEmembership{Senior Member,~IEEE}
\thanks{Hui-Chu Xiao, Wan-Lei Zhao, and Jie Lin are with Computer Science Department, Xiamen University, Xiamen, China. Wan-Lei Zhao is the corresponding authors.}
\thanks{Chong-Wah Ngo is with Computer Science Department, City University of Hong Kong. }
\thanks{Manuscript received xxx, xxx; revised xxx, xxx.}}

\maketitle

\begin{abstract}
The performance of instance search depends heavily on the ability to locate and describe a wide variety of object instances in a video/image collection. Due to the lack of proper mechanism in locating instances and deriving feature representation, instance search is generally only effective for retrieving instances of known object categories. In this paper, a simple but effective instance-level feature representation is presented. Different from other approaches, the issues in class-agnostic instance localization and distinctive feature representation are considered. The former is achieved by detecting salient instance regions from an image by a layer-wise back-propagation process. The back-propagation starts from the last convolution layer of a pre-trained CNN that is originally used for classification. The back-propagation proceeds layer-by-layer until it reaches the input layer. This allows the salient instance regions in the input image from both known and unknown categories to be activated. Each activated salient region covers the full or more usually a major range of an instance. The distinctive feature representation is produced by average-pooling on the feature map of certain layer with the detected instance region. Experiments show that such kind of feature representation demonstrates considerably better performance over most of the existing approaches. In addition, we show that the proposed feature descriptor is also suitable for content-based image search.
\end{abstract}

\begin{IEEEkeywords}
instance search, back-propagation, response peak, instance-level.
\end{IEEEkeywords}

\section{Introduction}

Different from image search, instance search is to hunt for images with the same object instances as a query image. The query instance is usually specified by a bounding box within an image or a video frame. To provide the evidence of search result, the location where a visual instance resides in a retrieved image should be presented for inspection. Instance search is widely used in various multimedia applications. In video editing, instance search serves as a function to return all the spatiotemporal locations of a query object instance, such as character, in a full-length video. In online survey, instance search is deployed to estimate the popularity of a brand (\textit{e.g.}, ``Coca cola'') by counting its appearance frequency over a large pool of Internet images. In online shopping, instance search enables fine-grained retrieval of product instances specific to a brand and style that a customer requests.

In instance search, the relevancy is grounded on the existence of instance rather than visual similarity of the whole image. Therefore, the conventional content-based image retrieval approaches that capture the global visual distribution of an image fall short for this problem. Typically, these approaches collapse features of different image regions into an embedded vector for retrieval. The visual characteristics unique to an instance may have been smoothed out during the embedding. As a consequence, the global feature is no longer distinctive for identification of individual instance, not mentioning the localization of instance as evidences of search result. The problem not only persists in hand-crafted visual features such as GIST~\cite{gist09:jegou}, but also deep features globally extracted from various neural networks~\cite{babenko2014neural, arxiv15:sharif}.

When instance search was first addressed in~\cite{Awad17}, the problem was coined as a sub-image retrieval task. Hand-crafted features such as SIFT~\cite{lowe2004distinctive} and SURF~\cite{bay2006surf} that are superior in local image matching were de-facto descriptors at that time. Although encouraging results are reported~\cite{Awad17}, these approaches are known to be limited to match textureless image patches and instances undergone non-rigid motions. While most of the descriptors are capable of generating thousands of local features from an image for matching, these features are extracted from regions rich of textures or corners. As a result, the object instances with textureless regions are under-represented. In addition to being invariant to 2D geometric transformations, local features can only tolerate certain degree of viewpoint and lighting changes. Particularly, the features are vulnerable to non-rigid deformations, which are widely observed in the real scenarios.

Recently, due to the great success of convolutional neural networks (CNNs) in learning high-level semantic features for image classification~\cite{alexnet}, object detection~\cite{ren2015faster} and instance segmentation ~\cite{he2017mask,li2017fully}, CNNs have been introduced to instance search~\cite{zhan2018instance}.
Using Fast R-CNN~\cite{girshick2015fast} as example, the instance-wise vector representation is produced through RoI-pooling from the region of feature maps corresponding to a  candidate object bounding box. The feature captures textureless region and is relatively robust to object deformation, if compared with global and local features. Despite satisfactory performance in instance search as reported in~\cite{zhan2018instance}, the main drawback of CNN-based solutions is their stringent demand on training data. In~\cite{zhan2018instance}, for example, pixel-wise annotation of instance location is required. The annotation effort is cost expensive and labor intensive. Furthermore, the learning process makes the CNNs more sensitive to object instances of known categories, by treating unseen categories as background class~\cite{he2017mask}. As a result, approach in~\cite{zhan2018instance} is only able to deal effectively with instances belonging to known object categories. This problem remains unaddressed if one switches to relying on  CNN-based object detection framework~\cite{ren2015faster,girshick2014rich, redmon2018yolov3} for instance-level feature extraction, such as~\cite{salvador2016faster}.

\begin{figure}[t]
\begin{center}
\subfigure[]{
	\label{fig:origin}
	\includegraphics[width=0.3\linewidth]{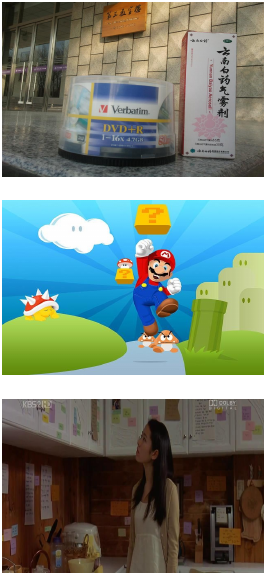}
}\hspace{-0.1cm}
\subfigure[]{
	\label{fig:boxes}
	\includegraphics[width=0.3\linewidth]{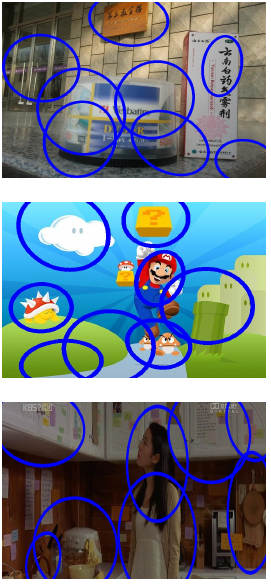}
}\hspace{-0.1cm}
\subfigure[]{
	\label{fig:heatmap}
	\includegraphics[width=0.3\linewidth]{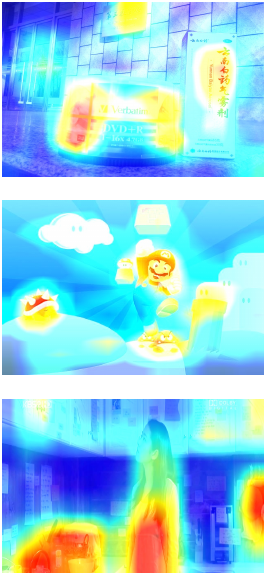}
}\hspace{-0.1cm}
\end{center}
   \caption{Instance localization: (a) original image,  (b) salient regions of instances being detected, (c) mean activation map as heat map overlapping on the original image. The heat map is color coded with red-tone indicates high response and blue-tone indicates low response. Each localized instance region is regularized by an estimated ellipse. }
\label{fig:dasrdemo}
\end{figure}

This paper targets for class-agnostic feature representation and localization for instance search. Leveraging on the pre-trained CNNs for visual classification, a new approach is proposed to detect the potential instances in an image. Starting from the last convolution layer, our approach detects response peaks and back-propagates them layer-by-layer. Those peaks support classification decision and generally refer to the regions residing on a visual instance. Through back-propagation, the effective receptive size of a salient region that corresponds to an instance and supports classification is activated at each layer. When reaching the outermost layer, \textit{i.e.}, the input image, the locations where the salient regions of instances reside can be uncovered. As the back-propagation starts from the last convolution layer rather than the prediction layer, the uncovered instances are class-agnostic and not restricted to the known categories. Figure~\ref{fig:dasrdemo} shows the examples of instance regions being detected, which are correspondent to the response peaks highlighted by the heat maps. A descriptor is then proposed to extract the feature of a salient region, by average-pooling over the feature maps corresponding to the region location.

The remainder of this paper is organized as follows. Section~\ref{sec:rela} reviews the state-of-the-art works in instance search and weakly supervised object detection. Our instance-level feature, namely deeply activated salient region (DASR), is presented in Section~\ref{sec:mthd}. The effectiveness of the proposed new feature representation is studied on the instance and image search in Section~\ref{sec:exp}. Section~\ref{sec:conc} concludes the paper.

\section{Related Work}
\label{sec:rela}
\subsection{ Instance Search} 
Instance search was addressed as a sub-image retrieval task before CNNs are introduced~\cite{Awad17} to visual object detection. The main image features being employed for this task are hand-crafted local descriptors such as SIFT and SURF. Through matching of local features, the instances relevant to a query are discovered in the candidate images for similarity search. Due to the high computational cost of direct point-to-point matching, the encoding approaches such as BoVW~\cite{Sivic03} and VLAD~\cite{jegou2011aggregating} are introduced to speed-up the matching. This line of approaches suffers from several limitations. First, non-rigid objects cannot be effectively handled~\cite{zhan2018instance}. Local features are vulnerable to non-rigid deformations and heavy viewpoint changes. Second, there is no guarantee that a feature being extracted will be unique to a particular instance. Instead, the features are often polluted by background or nearby objects of an instance. In most of the descriptors, the features are mostly extracted from local image patches located along the boundaries or corners of an object. When object instances are clustered in proximity, the image patch where a feature is derived from can occupy the partial regions of multiple instances. The problem also exists in local features extracted from deep neural network~\cite{noh2017large, paulin2015local}. Third, matching hundreds to even thousands of local features across two images is computationally prohibitive. Although matching can be considerably sped up by BoVW or VLAD representation, the search quality is also inevitably degraded due to vector quantization error.

Due to the satisfactory performance in image classification, pre-trained CNNs on classification tasks have been introduced to instance search. With the feature maps obtained by pre-trained models, R-MAC (regional maximum activation of convolutions)~\cite{tolias2015particular} aggregates features from several local regions into a global feature. Although encouraging results are obtained on image retrieval tasks, global features are infeasible for instance search. Weight aggregation are employed by CroW~\cite{kalantidis2016cross}, CAM~\cite{Jimenez_2017_BMVC}, BLCF-SALGAN~\cite{mohedano2018saliency} and Regional Attention~\cite{ kim2018regional} to address this problem. Region-level feature weighting allows the matching between global features to reflect the similarity between embedded instance features. The key idea is to  assign weights to different channels or different regions in the feature map during the feature pooling. The channels or regions which contribute more to the classification decision are assigned with higher weights. Due to the weighting scheme, the  instance which dominates in the image is highlighted in the embedded feature vector. While the scheme enables effective instance search, localization of query instance in the candidate images is not possible.

Recently, several attempts are devoted to instance-wise feature representation. The works rely on the fine-tuned CNNs that are designed for object detection or instance segmentation tasks. DeepVision~\cite{salvador2016faster} extracts region-level features from the bounding boxes generated by object detection framework. Due to the high computation cost, the features are only leveraged to rerank images at the top of a rank list. FCIS+XD~\cite{zhan2018instance}, instead, pixel-wisely extracts instance-level features from the instance segmentation map of fully convolutional network (FCN). The instances being considered are restricted to a limited number of object categories. PCL*+SPN~\cite{lin2019instance} extracts features from the object detection framework trained with image-level features. Despite leveraging on weakly supervised trained network, similar retrieval performance as FCIS+XD is reported in~\cite{lin2019instance}. The pitfall of this approach is that the network requires extra training stage and its disciminativeness to the unknown categories is undermined due to the extra training.

\subsection{Weakly Supervised Object Localization}
In a nutshell, robust instance-level feature representation relies on the ability to locate a wide variety of object instances. Compared to the fully supervised CNNs, weakly supervised networks that require only image-level labels for instance localization are more capable of dealing with larger number of object categories. Specifically, object regions are automatically influenced rather than manually provided during network learning. The existing approaches include proposal clustering learning (PCL)~\cite{pcl18:tang}, multiple instance learning (MIL)~\cite{maron1998framework, wan2019c, tang2017multiple} and weakly supervised instance segmentation (WSIS)~\cite{zhou2018weakly}. In PCL, a group of proposals are produced surrounding the regions that contribute to the classification score of one category. The proposals are reduced to several cluster centers during the learning, each of which is expected to cover a latent object of that category. In MIL, an image is viewed as a bag of object proposals. One object proposal is potentially a visual instance. During the training, MIL iteratively selects the instance with the highest confidence score until all the latent instances are detected. WSIS leverages the instance-level visual cue inside class activation maps (CAMs)~\cite{zhou2016learning}, which are produced by back-propagating iteratively the class-aware response peaks~\cite{zhou2018weakly}. The instance-aware cues are combined with class-aware cues and spatial continuity priors to produce instance segmentation masks. The best instance mask is selected after non-maximum suppression for each latent instance. Different from~\cite{zhou2018weakly}, the activation map in~\cite{wei2017selective} is produced by a simple aggregation of all the feature maps from $Pool_5$ of VGG-16~\cite{simonyan2014very}.  The regions whose activation values are higher than a threshold are detected as a part of latent instance. The detected neighboring regions are combined as the main instance, from where the feature representation are derived. This resulting feature is applied in the fine-grained image retrieval. Since only one instance is detected from one image, the approach is only applicable for single object instance retrieval.

Similar as~\cite{zhang2018top}, this paper also performs object localizations based on pre-trained CNNs that are used for image classification. The instance regions are localized by identifying the regions with high response in the iteratively back-propagated activation map. However, our approach differs from the existing works in three major aspects. Firstly, our approach does not intend to localize the full range of an instance. Instead, only the instance regions with high response in the activation map are localized. A region usually corresponds to the major part of an instance to be investigated by a network for classification. Secondly, the back-propagation starts from the last convolution layer of a network, instead of the prediction layer. This makes the localization remain sensitive to regions from the unknown instance categories. Finally, since no class-aware response is considered in our approach, no fine-tune training is involved.

\section{Deeply Activated Instance Region Detection}
\label{sec:mthd}
As witnessed in several recent works~\cite{zhou2018weakly,zhang2018top,zeiler2014visualizing}, salient regions of visual instances contributes significantly to the prediction of CNN. This property has been originally leveraged to interpret the behavior of a convolution neural network~\cite{zeiler2014visualizing}. The salient regions, despite not enclosing the entire instances, have been explored in various ways. Examples include modeling the top-down attention of a CNN ~\cite{zhang2018top} and weakly-supervised instance segmentation by integration of salient regions and other visual cues~\cite{zhou2018weakly}. In all these works, the salient regions are detected by back-propagating response peaks located in the classification layer. The back-propagation is essentially driven by the known categories that produce high classification confidences. As a result, the salient regions of unknown categories are overlooked throughout the process. The mechanism is not appropriate for instance search, which targets for all instances beyond the known categories of a CNN. 

In this paper,  the back-propagation is designed  to start from the last convolution layer, specifically the layer prior to the classification layer. The local maximums in this layer are detected and back-propagated layer-by-layer until reaching the input layer. By this way, salient regions of both known and unknown categories, which are activated layer-wisely, can be seamlessly located on the original image. In this section, building upon off-the-shelf pre-trained CNN, we present an end-to-end instance-level feature extraction framework.

\subsection{Activated Region Localization}
\label{sec:arl}
The forward-pass of a pre-trained network $\mathcal{N}$  produces a series of feature maps for an image $\mathit{I}$. Denote $X\in\mathbf{R}^{W{\times}H{\times}C}$ as the feature maps of the last convolution layer, the activation map is defined as the mean feature map of $\mathit{X}$, namely $\overline{X}\in\mathbf{R}^{W{\times}H}$. The map, which can be easily obtained by taking the average of $\mathit{X}$ over $C$ channels, signals the presence of instances. The regions on where $\overline{X}$ exhibits high values give clue to the confidences and positions of object instances. In a typical CNN, the peaks on the feature maps are assembled to support the decision-making in the next classification layer. Notice that $\overline{X}$ is prior to the classification layer, even the responses from unknown categories are visible as they have not been suppressed by the classification layer. Under this observation, the local maximums are detected on $\overline{X}$ with a window of size $3{\times}3$. These local maximums are viewed as the response peaks that network $\mathcal{N}$ discovers on image $\mathit{I}$.  Denote the set of peaks as $\mathit{Q}$, where each peak $\mathit{q}$ is attached with x-y position and a response value as confidence score.

\begin{figure*}[t]
\begin{center}
\includegraphics[width=0.9\linewidth]{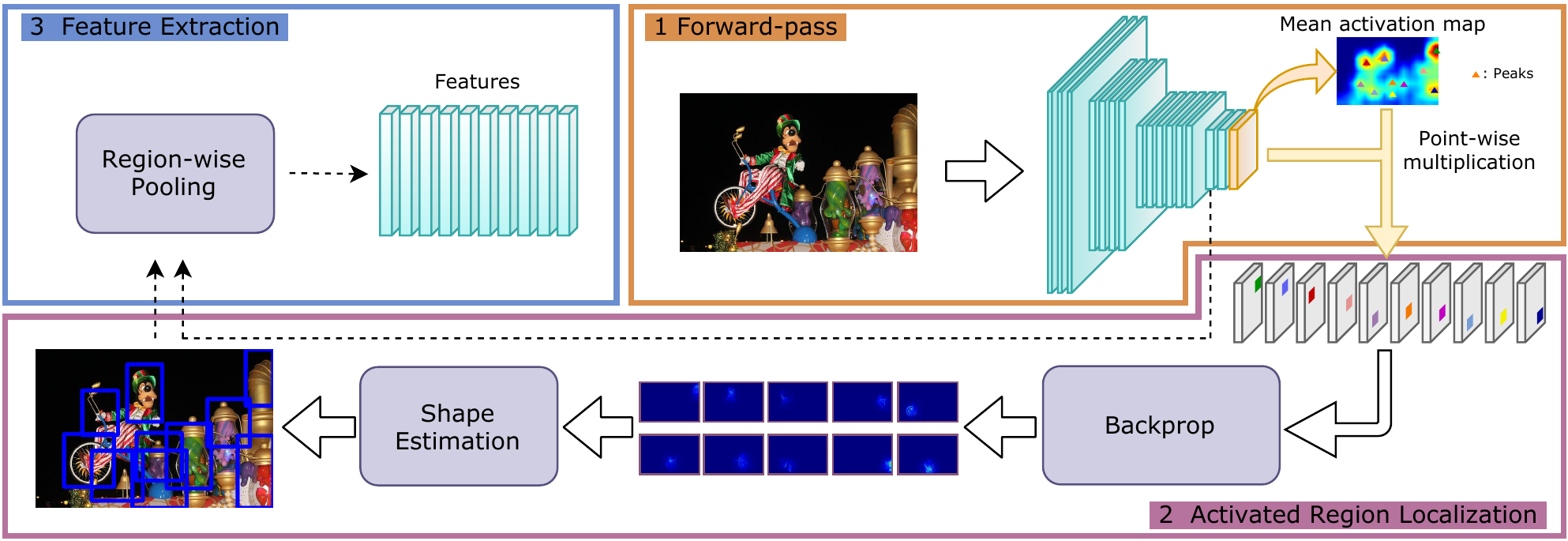}
\end{center}
   \caption{The pipeline of instance-level features extraction based on activated salient instance region. The mean activation map generated by a single forward-pass indicates the response of each potential instance region. The pattern localization process further localizes each salient region with a bounding box through a backprop and a shape estimation module. Final feature representations are built upon those localized boxes. Best viewed in color.}
\label{fig:pipeline}
\end{figure*}

Given the position of each peak $q \in Q$, a probability backprop process is adopted to locate the correspondent salient region on the input image. Following the similar process as~\cite{zhang2018top}, a top-down attention model is introduced to identify task-relevant input neurons that support the response peak \textit{q} in the last convolution layer.

Given no subsampling is performed in the convolution network, the convolution filter of one intermediate convolution layer is denoted as $F\in{\mathbf{R}^{W_f{\times}H_f{\times}C_{out}{\times}C_{in}}}$, where $W_f{\times}H_f$ is the spatial size of a filter.  $C_{in}$ and $C_{out}$ are the channel dimensions of input and output feature maps respectively. The input and output feature maps of this convolution layer are denoted as $A$ and $B$. The activation from each spatial location in $A$ and $B$ could be accessed by $A_{x,y}$ and $B_{i,j}$ respectively. The trained weights related to $A_{x,y}$ and $B_{i,j}$ are accessed with $F_{x-i,y-j}$. The feed-forward process to generate the output tensor $B$ is formulated as 
\begin{equation}
B_{i,j} = \sigma(\sum_{x=i-\frac{W_f}{2}}^{i+\frac{W_f}{2}}\sum_{y=j-\frac{H_f}{2}}^{j+\frac{H_f}{2}}F_{x-i,y-j}A_{x,y}+b),
\end{equation}
where $b$ is the bias of convolution layer and $\sigma$ represents the non-linear activation function. 

Now let's consider to back-propagate peak pixels in the last output layer $B$. Notice that only the peak positions that are detected from the last convolution layer are considered.  
Precisely, the idea is to identify the positions in $A$ that contribute to the score of response peak at $B_{i,j}$, \textit{i.e.}, \textit{q}. Following~\cite{zhou2018weakly,zhang2018top}, this issue is modeled as a prior probability distribution $P(A_{x,y})$ over output response. $B_{i,j}$ is assumed to be the only winner which takes responses from all the positions in $A$. Therefore, given $P(B_{i,j})$ and $P(A_{x,y}|B_{i,j})$ are known, we are able to work out $P(A_{x,y})$ , viz. the probability that the task-relevant neurons in $B$ come from $A_{x,y}$.

For computational convenience, $P(B_{i,j})$ is approximated by $B_{i,j}$ in the last convolution layer. As a consequence, the prior probability of input $A$ is given as
\begin{equation}
\label{eqn:pa}
P(A_{x,y}) = \sum_{i=x-\frac{W_f}{2}}^{x+\frac{W_f}{2}} \sum_{j=y-\frac{H_f}{2}}^{y+\frac{H_f}{2}} P(A_{x,y}|B_{i,j})P(B_{i,j}). 
\end{equation}
In Eqn.~\ref{eqn:pa}, the conditional probability $P(A_{x,y}|B_{i,j})$ is defined as
\begin{equation}
P(A_{x,y}|B_{i,j}) = 
\begin{cases}
Z_{i,j}A_{x,y}F_{x-i,y-j}, & \text{if } F_{x-i,y-j} > 0 \\
0, & otherwise. \\
\end{cases}  \label{eqn:pab}
\end{equation}
where $Z_{i,j}$ is a normalization factor to make sure that $\sum_{i,j}P(A_{x,y}|B_{i,j})=1$. The above conditional probability estimates the winning probability of position $(x,y)$ in $A$ given position $(i,j)$ in $B$ is a winning neuron. The estimation is affected by the activation $A_{x,y}$ and the value within convolution filter $F_{x-i,y-j}$ which relates to $A_{x,y}$ and $B_{i,j}$. 

With Eqn.~\ref{eqn:pa}, each position in $A$ is assigned with a probability weight. In the next round of back-propagation, the resulting $P(A_{x,y})$ becomes $P(B_{i,j})$, and $P(A_{x,y}|B_{i,j})$ can be easily estimated in the same manner with Eqn.~\ref{eqn:pab}.  

\begin{figure*}[t]
\begin{center}
\subfigure[Activated pixel region for each response peak]{
	\label{sfig:neuron}
    \includegraphics[width=1\linewidth]{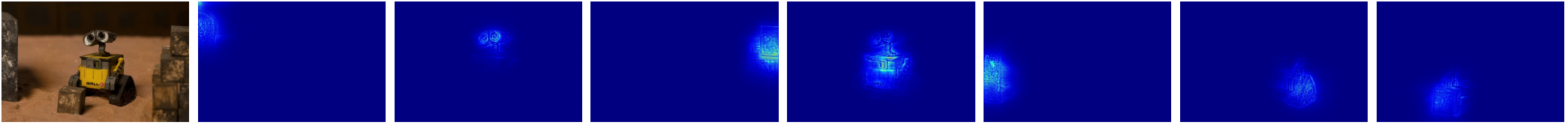}
}
\subfigure[Ellipse estimation in each activated region]{
	\label{sfig:ellips}
    \includegraphics[width=1\linewidth]{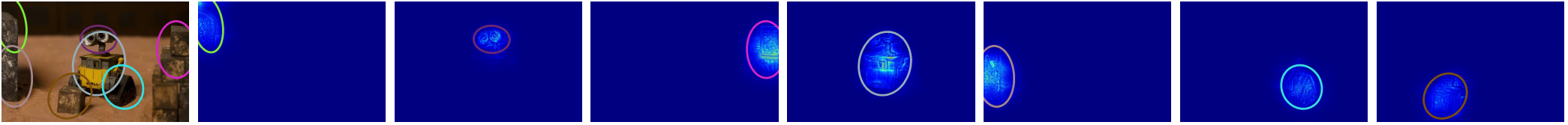}
}
\end{center}
   \caption{The illustration of deeply activated salient regions in an input image and ellipse estimation on each activated region. The first row shows the input image and activated regions with all the detected  seven peaks. The corresponding estimated ellipse for each activated region is shown on the second row.}
\label{fig:ellipse}
\end{figure*}

In addition to convolution layers, the backprop process also passes through other intermediate layers, \textit{e.g.}, pooling layers. The average-pooling layers are regarded as performing an affine transformation on the response values of the input neurons~\cite{zhang2018top}. Therefore, the average-pooling layer is treated as a convolution layer that is performed within one-to-one feature map pair. For max-pooling layers, error back-propagation is adopted to perform backprop in~\cite{zhang2018top}. However, blanks are introduced for sub-sampled max-pooling layers. In order to avoid such blanks, the same backprop process as convolution layer is used for max-pooling layers within one-to-one feature map pair, with the weights of all-one values.

To this end, all types of layers that the back-propagation may pass through are appropriately considered with the same manner.  Eqn.~\ref{eqn:pa} applies to all the layers throughout the convolution network $\mathcal{N}$. Therefore, the back-propagation process proceeds layer by layer smoothly until it reaches the input layer. Finally, the probability that each pixel in image \textit{I} contributes to a final response peak $q$ is estimated. This leads to a probability map $M$, which is in the same size as image \textit{I}, for one response peak $q$. The probabilities in $M$ are normalized to the range $[0,1]$. 

Values in $M$ indicate the degree that corresponding pixels contribute to peak $q$. Due to the large receptive field of the last convolution layer, pixels which do not contribute to the response peak are still assigned with low probabilities. As a result, the activated region is usually larger than it is supposed to be. A threshold $\tau$ is introduced to filter out pixels with little contribution. In the paper, $\tau$ is fixed to \textit{0.1}. As shown in Figure~\ref{fig:ellipse}, the activated pixels in general concentrate on a local region, which basically implies a potential instance in the image. With all the pixels $r(x,y)$ in $M$ that are greater than $\tau$, this activated local region is approximated by an ellipse. The parameters of the ellipse are regularized by the second moment matrix derived from all pixels $r(x,y)$
\begin{equation}
\sum_{r(x, y) \ge \tau} \left[ \begin{array}{cc}
x^2 & x{\cdot}y \\
x{\cdot}y & y^2
\end{array}
\right].
\label{eqn:region}
\end{equation}
Figure~\ref{fig:ellipse}(a) illustrates the probability maps produced from seven response peaks in one image. The corresponding shape estimation results are shown in the second row of Figure~\ref{fig:ellipse}. As shown in the figure, each detected region corresponds to one salient region in the image. It could cover an entire instance or a major salient region of an instance. The final localization bounding box is the circumscribed rectangle of the estimated ellipse. The feature used to describe this detected region could be derived from the corresponding area of a feature map. Since this feature is produced by activating the salient region via a deep convolution network, it is called \textit{deeply activated salient region} (\textbf{DASR}) from now on. 

\subsection{Enhanced Instance Region Detector}
\label{sec:nms}
In the above activation process, only the pixels with peak response in the last convolution layer are back-propagated. In practice, it is possible that more than one instance share one peak response as they are close to each other. In this case, a detected salient region will only cover one of the instances. The other neighboring instances are over-shadowed. To alleviate this issue, we consider to back-propagate more number of pixels in $\overline{X}$. Specifically, all the pixels whose response is higher than the average value of $\overline{X}$ are back-propagated one by one.  As a result, more number of salient regions are produced. However, it is possible that two salient regions overlap with each other and cover over the same instance. In order to reduce the representation redundancy and select out the most salient regions, non-maximum suppression (NMS) is employed as \cite{zhou2018weakly}.

The NMS is operated as follows. The intersection-over-union (IoU) threshold of NMS is given as $\beta$. Each candidate salient region is attached with a corresponding response value in $\overline{X}$.  NMS starts by selecting the salient regions with the highest score uniformly across the space of $\overline{X}$. The remaining regions are screened by comparing their IoU with the set of selected salient regions. Specifically, a region is discarded if its IoU with one of the already selected regions is greater than $\beta$. The valid setting for parameter $\beta$ is further studied in the experiment section.

With the new detection procedure, salient regions which attain the highest response in a local are kept. While the regions from other potential instances, which have been over-shadowed before, could be activated as long as their overlapping with the most salient region in the local is below a threshold. On average, \textit{12} regions (in contrast to 7 regions before)\footnote{Statistics are made on \textit{1} million images crawled from Flickr.} are detected in one image after MNS when $\beta$ is set to \textit{0.3}. This enhanced detector is named as DASR*. Its effectiveness is further verified in the experiment section.  

\textbf{Discussion} Note that the proposed back-propagation can start from the response peaks of any convolution layer. The peaks at a shallow layer correspond to regions with more fine-grained local patterns. In our case, the aim is to discover latent instances. The last convolution layer is the one that directly supports the classification decision. A high response peak in this layer is an integral of visual clues from one instance of known or unknown category. One or several response peaks of one category from this layer are further integrated by the next layer to make a classification decision. It is therefore  appropriate to choose the last convolution layer in our case. However, it is open to select other layers when the task changes. Our approach is not restricted to image feature extraction, it is a generic feature extraction pipeline in the sense that it is feasible as long as CNNs are pre-trained for classification.

\subsection{Feature Description}
\label{sec:feat}
\textbf{Instance level.} The descriptor of a salient region can be extracted by max or average pooling over the feature maps of its corresponding location. The feature descriptor will be compact and uniform in length. In our paper, average-pooling is selected over max-pooling for its consistently better performance. Theoretically speaking, feature map from any layer could be used to derive the feature descriptor. However, the distinctiveness varies from layer to layer. For instance, we find that feature derived from ``Block4'' in ResNet-50~\cite{he2016deep} shows considerably better performance over other layers across different datasets. The details will be followed up in the ablation study. The generated features are first $\textit{l}_2$-normalization, and then undergone PCA whitening before the second round of $\textit{l}_2$-normalization. We call the instance-level feature as DASR descriptor.

\textbf{Image level.} A salient region covers either the entire instance or a semantic part of an instance. Instead of matching every instance or part locally across images for search, an image can be represented as a bag-of-instances for global similarity search. To achieve that, DASR descriptors that are extracted from each image are embedded into one vector with VLAD. Small visual word vocabulary of DASR is trained in advance. To this end, a collection of DASR features are converted into a long dense vector. The schemes proposed in VLAD*~\cite{mm13:vlad} are adopted to boost the performance. As revealed in the experiment, encouraging performance is reported on image search task. It outperforms classic hand-crafted descriptor SIFT and shows competitive performance with the existing deep features that are designed specifically for image search.

\section{Experiments}
\label{sec:exp}
\subsection{Datasets and Experimental Setup}
The proposed instance-level feature DASR is evaluated in two search tasks, namely instance search and conventional content-based image retrieval. Instance search is conducted on three benchmarked datasets: Instance-160~\cite{zhan2018instance}, Instance-335 and INSTRE~\cite{wang2015instre}. Instance-160 and Instance-335 datasets are derived from the video sequences originally used for single visual object tracking evaluation. In Instance-160, there are \textit{160} queries and \textit{11,885 }reference images. The query instances belong to \textit{80} object categories labelled in Microsoft COCO dataset~\cite{lin2014microsoft}. In order to test the scalability of the proposed instance-level feature, Instance-160 is augmented with \textit{175} extra queries that are havested from GOT-10K~\cite{huang2018got}, Youtube BoundingBoxes~\cite{real2017youtube} and LaSOT~\cite{fan2019lasot}. These video datasets are originally designed for object tracking evaluation. These newly added \textit{175} query instances are mostly out of the coverage of Microsoft COCO \textit{80} categories and the backgrounds are under severe variations. This leads to an augmented evaluation dataset Instance-335.  In this dataset, there are \textit{335} queries and \textit{40,914} reference images. For INSTRE dataset, there are \textit{27,293} images in total. Following with the evaluation protocal in \cite{iscen2017efficient}, \textit{1,250} images\footnote{One query is selected from one image.} are treated as the queries, leaving the remaining \textit{27,293} images as references. For all the three datasets, the bounding boxes are provided both in the query and relevant reference images.

For image retrieval task, DASR is evaluated on three popular evaluation benchmarks: Holidays~\cite{jegou2008hamming}, Oxford Buildings (Oxford5k)~\cite{philbin2007object}, and Paris (Paris6k)~\cite{philbin2008lost}. In our implementation, the images of these datasets are re-sized to \textit{512} pixels on the long side, while preserving the aspect ratio of original images. Following the convention in the literature, the search performance of the two tasks is measured with mean Average Precision (mAP). For Instance-160 and Instance-335, the search performance is evaluated with varying top-\textit{k}, where \textit{k} varies from \textit{50} to \textit{100}. This is because the number of true-positives for each instance query varies from several to a few hundred for both Instance-160 and Instance-335. 

The proposed feature extraction can be carried out using any CNN classification networks. Here, we report experimental results based on ResNet-50 and VGG-16, which are widely used by different applications. As revealed in the later experiment, the performance using ResNet-50 is considerably higher. Hence, most of the presented results will be based on ResNet-50 by default unless otherwise stated. The feature extraction is implemented under Tensorflow framework. Experiments are run on an Nvidia GTX 1080 Ti. 


In the first experiment, an ablation study is presented to investigate the suitablity of feature maps at different layers for feature extraction. In addition, we also verify the parameter setting in NMS, \textit{i.e.}, the IoU rate $\beta$ for pruning instance candidates. To this end, two groups of comparative studies are presented. First, the performance of DASR is studied in comparison to R-MAC~\cite{tolias2015particular}, CroW~\cite{kalantidis2016cross}, CAM~\cite{Jimenez_2017_BMVC}, BLCF~\cite{mohedano2018saliency}, BLCF-SalGAN~\cite{mohedano2018saliency}, Regional Attention~\cite{kim2018regional}, DeepVision~\cite{salvador2016faster}, FCIS+XD~\cite{zhan2018instance} and PCL*+SPN~\cite{lin2019instance} in the instance search task. Note that the comparison is based on instance-level matching. Next, we assess the performance of converting the instance features into a global image feature for content-based retrieval. The comparison is made against BoVW+He~\cite{jegou2008hamming}, SIFT+VLAD*~\cite{mm13:vlad}, R-MAC, CroW, CAM, BLCF, BLCF-SalGAN, Regional Attention and DeepVision.

\begin{figure}[t]
\begin{center}
    \includegraphics[width=0.9\linewidth]{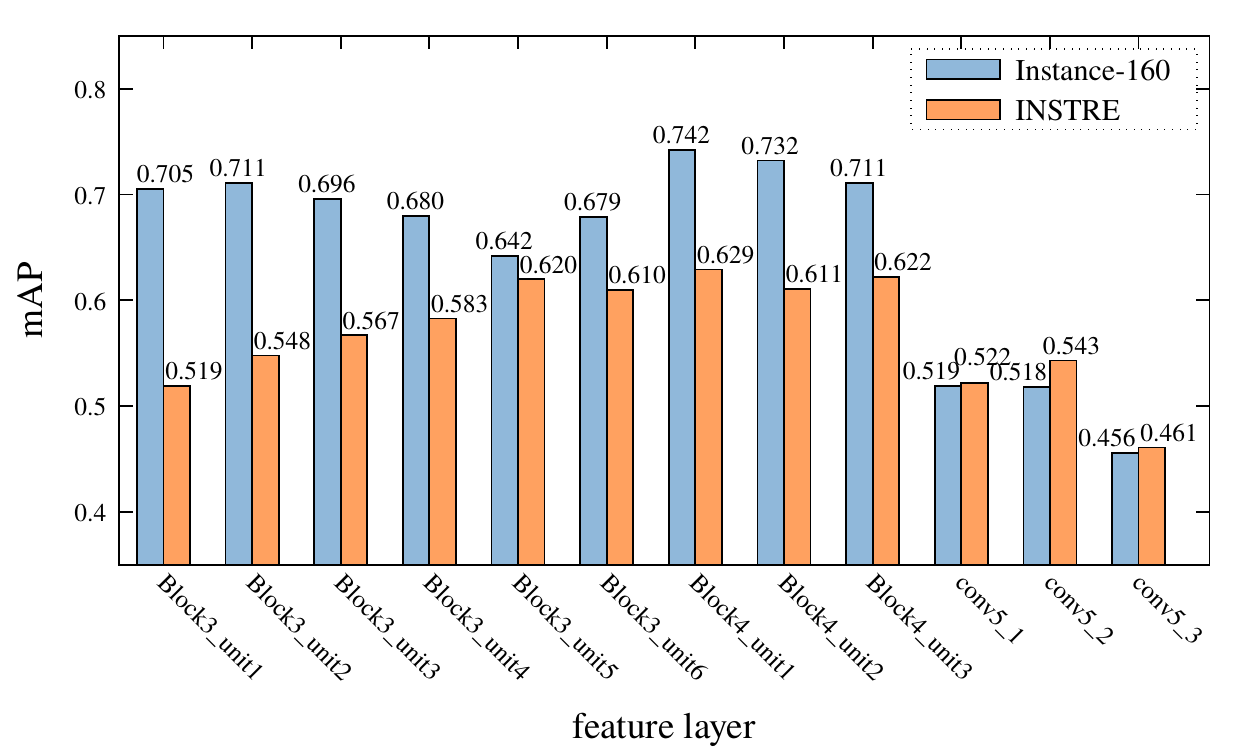}
\end{center}
   \caption{Performance of DASR on Instance-160 and INSTRE datasets with features derived from different convolutional layers of ResNet-50 and VGG-16.}
\label{fig:ftmap}
\end{figure}

\subsection{Ablation Study}

\subsubsection{Feature Selection}
\label{sec:selft}
Given the detected salient region, feature map from each convolutional layer could be used to derive the feature descriptor. Nevertheless, it has been widely witnessed that the search performance varies across different layers \cite{babenko2014neural, zhan2018instance, lin2019instance}. For this reason, ablation analysis is conducted to seek for the best suitable layer for instance search. Layers from the last two blocks of ResNet-50, namely Block3 and Block4, are investigated since deeper layers are observed to contain semantic-level information. Following the original implementation of ResNet-50, six and three bottlenecks are built within Block3 and Block4 respectively. The output feature maps from above \textit{9} bottlenecks are respectively used to derive features for DASR regions. The first bottleneck in Block3 is given as Block3\_unit1, and the rest are denoted in the same manner. 
For VGG-16 network,  the back-propagation starts from the feature map of the \textit{5}th pooling layer, which is the last layer prior to the fully connected layers. Features are extracted from feature maps of the three convolutional layers on the \textit{5}th stage. They are given as conv5\_1, conv5\_2 and conv5\_3 respectively. In this experiment, no NMS is adopted in the region detector.

Figure~\ref{fig:ftmap} shows the performance of DASR on Instance-160 and INSTRE with features output from different layers of two backbones. A wide performance gap is observed between the networks. The gap is due to the difference in encoded patterns and feature discriminability between the networks. The result of instance localization is directly influenced by the encoded patterns. In general, the regions derived via ResNet-50 shows high localization accuracy. Moreover, feature maps from ResNet-50 are more discriminative than those of VGG-16, which is in line with the observations from many other works. Overall, features derived from Block4\_unit1 show  the best performance on both datasets. As a result,  it is selected as the default configuration in the rest of experiments.

\subsubsection{Configurations on DASR*}
\begin{figure}[t]
\begin{center}
    \includegraphics[width=0.75\linewidth]{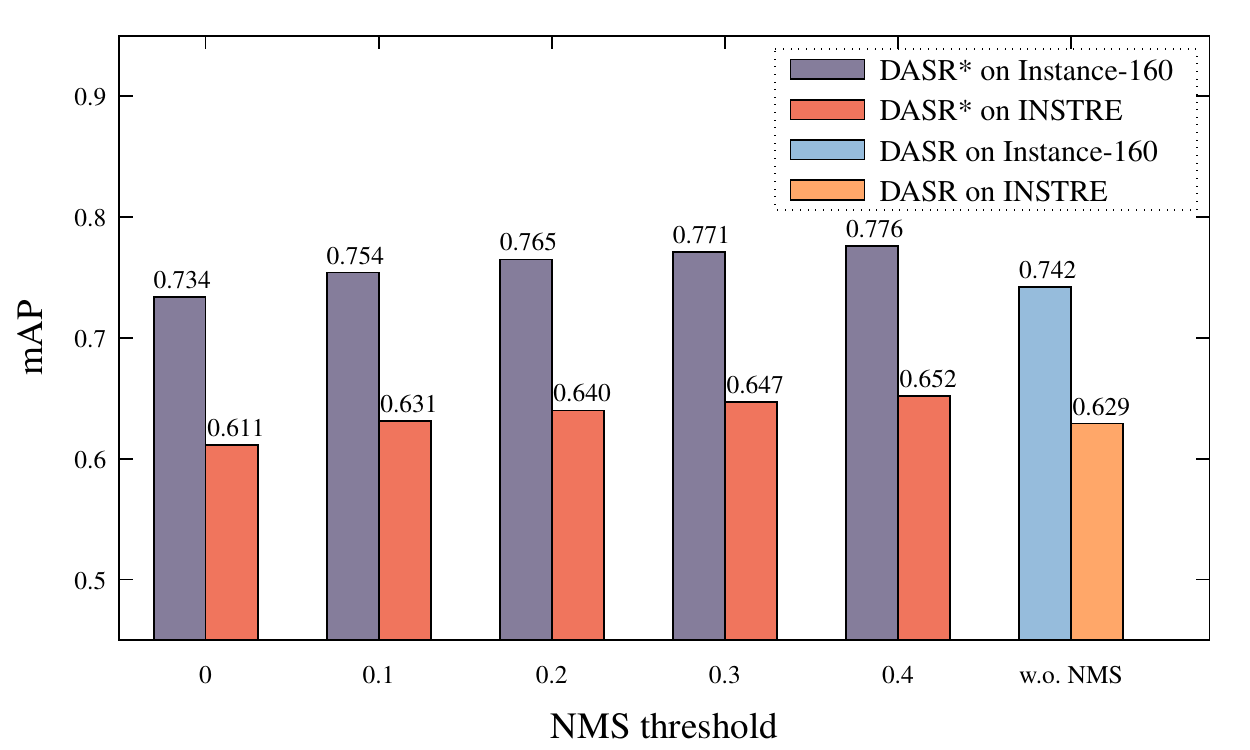}
\end{center}
   \caption{The performance of DASR on Instance-160 and INSTRE datasets with different NMS threshold $\beta$s and wihtout NMS.}
\label{fig:nmsrslt}
\end{figure}
In the second study, we further investigate the effectiveness of the enhanced detector DASR* and the appropriate  setting for overlapping rate parameter $\beta$ in NMS. In this study, the enhanced detection procedure presented in Section~\ref{sec:nms} is performed on ResNet-50. Performance with different settings of $\beta$ is presented in Figure~\ref{fig:nmsrslt}. The performance is also compared to the one without NMS. As shown in the figure, DASR* outperforms DASR when the overlapping rate is higher than \textit{0.1}. Moreover, the larger overlapping rate $\beta$ leads to better performance, since more salient regions are kept for one image. The highest performance is attained when $\beta=0.4$, which also leads to much more number of detected regions. Specifically, the number of detected regions is roughly doubled over the case of being without NMS. As a trade-off between performance and computational cost, $\beta$ is set to \textit{0.3} in the rest of our experiments.


\subsection{Instance Search}
\begin{figure*}[t]
\begin{center}
\includegraphics[width=1\linewidth]{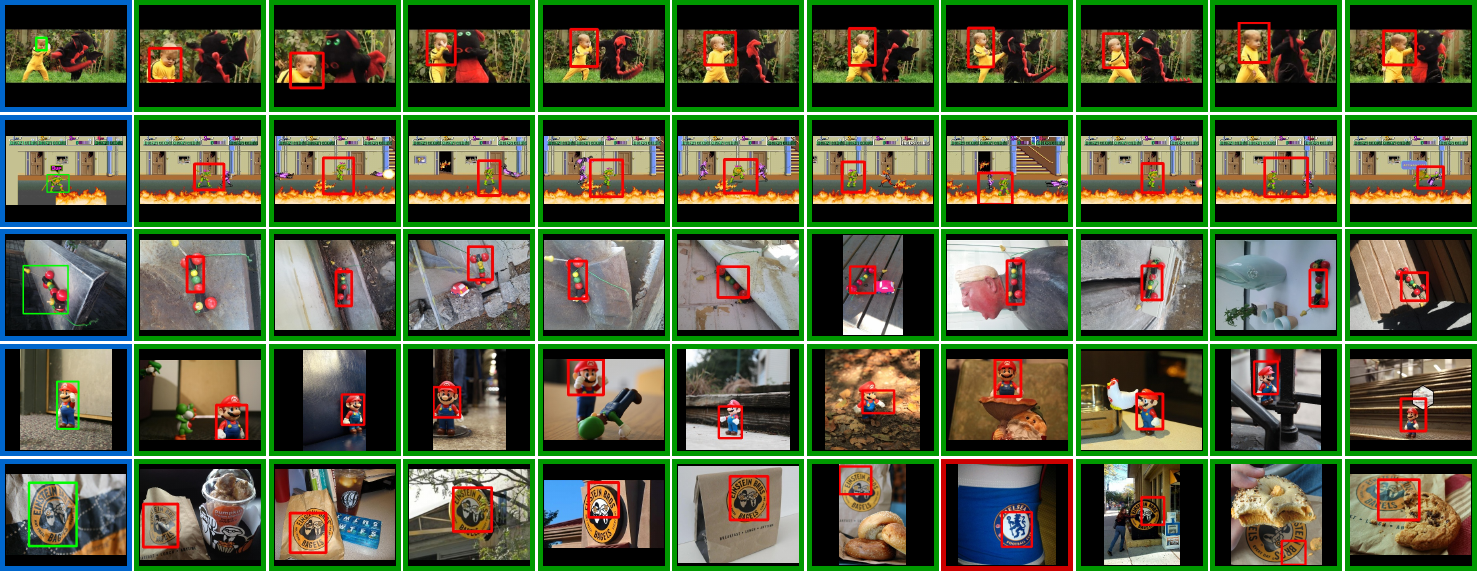}
\end{center}
   \caption{Examples of top-10 retrieved instances by five query examples in Instance-335 (the first two rows) and INSTRE (the last three rows). The left most column shows the queries, while the remaining columns display the retrieved images sorted in descending order. The true and false positive images are enclosed by green and red borders respectively.}
\label{fig:top10}
\end{figure*}
\subsubsection{Comparison to State-of-the-Art Approaches}
DASR is compared against several representative approaches in the literature. The approaches are categorized accordingly to the degree of supervision involved to train a network for instance search. The first group of approaches capitalize on the convolutional features derived from pre-trained CNNs without model fine-tuning. These approaches include R-MAC~\cite{tolias2015particular}, CroW~\cite{kalantidis2016cross}, CAM~\cite{Jimenez_2017_BMVC}, BLCF~\cite{mohedano2018saliency}, BLCF-SalGAN~\cite{mohedano2018saliency}, and Regional Attention~\cite{kim2018regional}. In contrast, the second group of approaches fine-tune the pre-trained CNN with extra training examples in COCO dataset. The only approach falls in this group is PCL*+SPN~\cite{lin2019instance}. The last group of approaches are DeepVision~\cite{salvador2016faster} and FCIS+XD~\cite{zhan2018instance}, which leverages on R-CNN and FCN respectively to extract features. Same as the second group, the object detection models are also fine-tuned with the training data in COCO dataset. Additionally, object-level labels are required. For instance, DeepVision is trained with object bounding boxes while FCIS+XD requires the instance masks of objects. Note that DeepVision also adopts re-ranking and query expansion strategies to improve the search performance.  
During retrieval, the first group of approaches collapses all the features into one vector for retrieval. The second and third groups, instead, treat each instance individually as a retrieval unit. Specifically, all instances of an image are compared to the query instance, and the similarity is set equal to the instance with the highest matching score. DASR and DASR*, similar to the first group, requires only pre-trained CNN. On the other hand, as the second and third groups, the extracted instances from an image are treated independently during retrieval. For convenience, we name the three groups of approaches as ``pre-trained'', ``weak'' and ``strong'' respectively.

\begin{table}
\begin{center}
\caption{Performance comparison on Instance-160 and Instance-335}
\label{tab:ints335}
\resizebox{0.5\textwidth}{!}{
\subtable[Instance-160]{
\begin{tabular}{|p{0.15\textwidth}|p{0.09\textwidth}<{\centering}|p{0.04\textwidth}<{\centering}|p{0.045\textwidth}<{\centering}|p{0.053\textwidth}<{\centering}|p{0.04\textwidth}<{\centering}|}
\hline
Approach & Model-Type & Dim. & Top-50 & Top-100 & All \\
\hline\hline
R-MAC~\cite{tolias2015particular} & pre-trained & 512 & 0.268 & 0.307 & 0.358 \\
CroW~\cite{kalantidis2016cross} & pre-trained & 512 & 0.239 & 0.284 & 0.338 \\
CAM~\cite{Jimenez_2017_BMVC} & pre-trained & 512 & 0.256 & 0.302 & 0.358 \\
BLCF~\cite{mohedano2018saliency} & pre-trained & 336 & 0.487 & 0.592 & 0.653 \\
BLCF-SalGAN~\cite{mohedano2018saliency} & pre-trained & 336 & 0.493 & 0.596 & 0.656 \\
Regional Attention~\cite{kim2018regional} & pre-trained & 2,048 & 0.318 & 0.389 & 0.459 \\
DeepVision~\cite{salvador2016faster} & strong & 512 & 0.541 & 0.666 & 0.731 \\
FCIS+XD~\cite{zhan2018instance}$^{\ddag}$ & strong & 1,536 & 0.575 & 0.659 & 0.724 \\
PCL*+SPN~\cite{lin2019instance} & weak & 1,024 & 0.583 & 0.661 & 0.724 \\ \hline
DASR & pre-trained  & 2,048 & 0.591 & 0.680 & 0.742 \\
DASR* & pre-trained  & 2,048 & \textbf{0.614} & \textbf{0.711} & \textbf{0.771} \\
\hline
\end{tabular}}}
\resizebox{0.5\textwidth}{!}{
\subtable[Instance-335]{
\begin{tabular}{|p{0.15\textwidth}|p{0.09\textwidth}<{\centering}|p{0.04\textwidth}<{\centering}|p{0.045\textwidth}<{\centering}|p{0.053\textwidth}<{\centering}|p{0.04\textwidth}<{\centering}|}
\hline
R-MAC~\cite{tolias2015particular} & pre-trained & 512 & 0.234 & 0.315 & 0.375 \\
CroW~\cite{kalantidis2016cross} & pre-trained & 512 & 0.159 & 0.225 & 0.321 \\
CAM~\cite{Jimenez_2017_BMVC} & pre-trained & 512 & 0.194 & 0.263 & 0.347 \\
BLCF~\cite{mohedano2018saliency} & pre-trained & 336 & 0.246 & 0.358 & 0.483 \\
BLCF-SalGAN~\cite{mohedano2018saliency} & pre-trained & 336 & 0.245 & 0.350 & 0.469 \\
Regional Attention~\cite{kim2018regional} & pre-trained & 2,048 & 0.242 & 0.351 & 0.488 \\
DeepVision~\cite{salvador2016faster} & strong & 512 & 0.402 & 0.521 & 0.620 \\
FCIS+XD~\cite{zhan2018instance} & strong & 1,536 & 0.403 & 0.500 & 0.593 \\
PCL*+SPN~\cite{lin2019instance} & weak & 1,024 & 0.380 & 0.475 & 0.580 \\ \hline
DASR & pre-trained & 2,048 & 0.419 & 0.558 & 0.699 \\
DASR* & pre-trained & 2,048 & \textbf{0.433} & \textbf{0.580} & \textbf{0.724} \\
\hline
\end{tabular}}}\\
$^{\ddag}$ digits are cited from the referred paper.
\end{center}
\end{table}

\begin{table}
\scriptsize
\begin{center}
\caption{Performance comparison on INSTRE}
\label{tab:instre}
\begin{tabular}{|l|c|c|c|}
\hline
Approach & Model-Type & Dim. & All \\
\hline\hline
R-MAC~\cite{mohedano2018saliency}$^{\ddag}$ & pre-trained & 512 & 0.523 \\
CroW~\cite{mohedano2018saliency}$^{\ddag}$ & pre-trained & 512 & 0.416 \\
CAM~\cite{Jimenez_2017_BMVC} & pre-trained & 512 & 0.320 \\
BLCF~\cite{mohedano2018saliency}$^{\ddag}$ & pre-trained & 336 & 0.636 \\
BLCF-SalGAN~\cite{mohedano2018saliency}$^{\ddag}$ & pre-trained & 336 & \textbf{0.698} \\
Regional Attention~\cite{kim2018regional} & pre-trained & 2,048 & 0.542 \\
DeepVision~\cite{salvador2016faster} & strong & 512 & 0.197 \\
FCIS+XD~\cite{zhan2018instance} & strong & 1,536 & 0.067 \\
PCL*+SPN~\cite{lin2019instance}$^{\ddag}$ & weak & 1,024 & 0.575 \\ \hline
DASR & pre-trained & 2,048 & 0.629 \\
DASR* & pre-trained & 2,048 & 0.647 \\
\hline
\end{tabular}\\
$^{\ddag}$ digits are cited from the referred paper.
\end{center}
\end{table}

Tab.~\ref{tab:ints335} and Tab.~\ref{tab:instre} show the performance of different approaches on three datasets. In general, all the approaches show steady performance degradation when being tested on more challenging dataset Instance-335. Among all the approaches, DeepVision, FCIS+XD, PCL*+SPN, and DASR which produce instance-level feature demonstrate better performance on all the datasets. The instance-level features are more robust to background variations, which has been well illustrated in~\cite{lin2019instance}. Among these instance-level features, DASR and DASR* show consistently satisfactory performance on both datasets. In contrast, the performance from FCIS+XD drops considerably on Instance-335 and INSTRE. It is simply because there are many instance categories outside the coverage of Microsoft COCO-80 on which FCIS+XD training fully relies.
Interestingly, DASR* even outperforms FCIS+XD considerably on Instance-160, where all the query instances are well trained in FCIS+XD. FCIS+XD is capable of generating more precise instance region. The reason that DASR* outperforms FCIS+XD mainly attributes to the better discriminativeness of the feature maps. Notice that the feature maps in FCIS+XD are trained for instance segmentation. It carries more localization information rather than semantic information of an instance. BLCF-SalGAN, although showing the overall best performance on INSTRE, is sensitive to various image transformations. This is evidenced in the datasets Instance-160 and Instance-335, where its performance is not satisfactory with the presence of non-rigid transformation.

\begin{figure*}[t]
\begin{center}
\subfigure[INSTRE]{
	\label{fig:iou1}
    \includegraphics[width=0.42\linewidth]{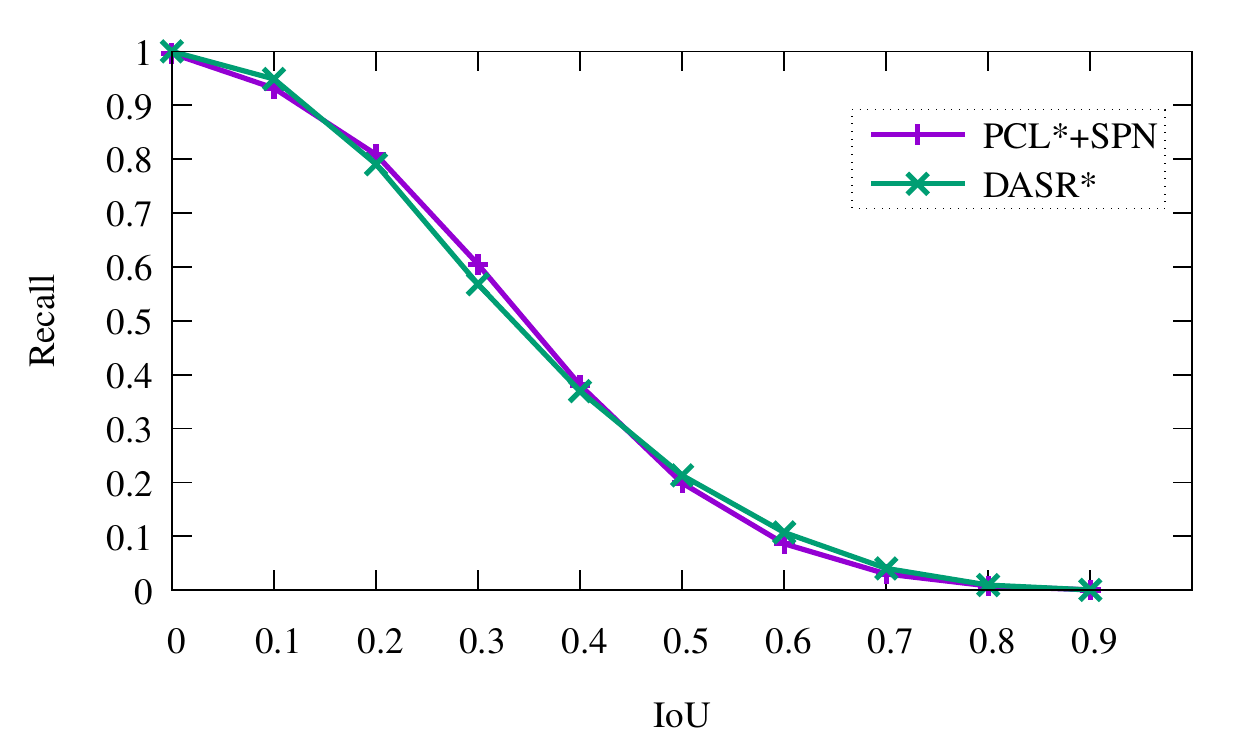}
}
\subfigure[Instance-335]{
	\label{fig:iou2}
   	\includegraphics[width=0.42\linewidth]{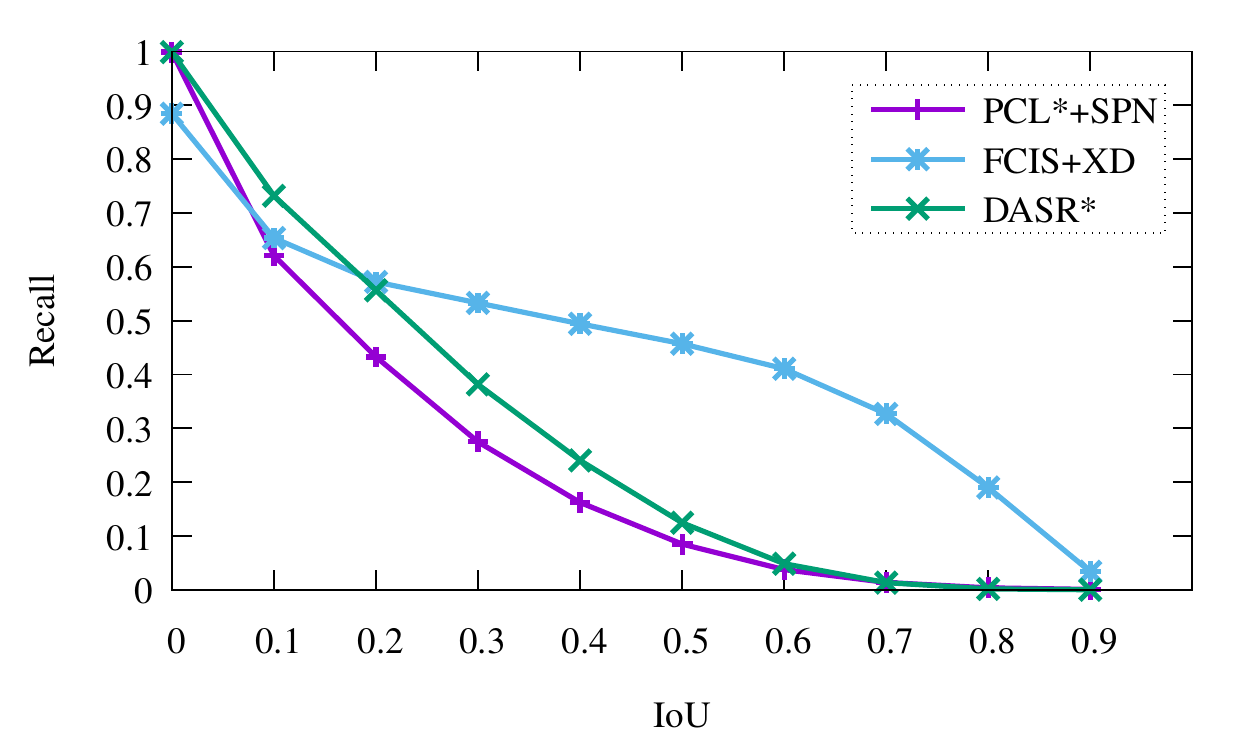}
}
\end{center}
   \caption{Recall-IoU curves on INSTRE and Instance-335.} 
\label{fig:iou}
\end{figure*}

Top-10 retrieval results from our approach are illustrated in Figure~\ref{fig:top10}. The first row is an example showing the robustness of DASR* to non-rigid transformation. The second row shows a cartoon character, which does not belong to any known categories in the COCO dataset, as the query instance. DASR* successfully retrieves and indicates the location of cartoon character for the top-10 ranked images. The result verifies that the proposed back-propagation mechanism manages to capture the instance of categories new to a pre-trained network. However, DASR* could be sensitive to instances with similar shape appearance but different fine-grained details. One example is shown in the last row, where a logo with different printing and icon than the query logo is retrieved.


\subsubsection{Instance Localization Accuracy}
In this experiment, we further study how well the detected regions overlaps with instances in the ground-truth. Following~\cite{hosang2015makes}, recall is adopted to measure the fraction of a detected region that overlaps with its ground-truth instance based on IoU threshold. Different values of recall with varying IoU thresholds are reported. The experiment compares DASR* with FCIS+XD and PCL*+SPN since these are the only approaches capable of locating object instances. Figure~\ref{fig:iou} shows the recall-IoU curves of different approaches on two datasets. The localization performance of FCIS+XD is not compared on INSTRE since a large portion of the instance categories are not covered in its training dataset. 

As shown in the Figure~\ref{fig:iou}(a), our approach shows similar or even slightly better performance over PCL*+SPN, which is designed to localize the whole instance from images. While as shown in Figure~\ref{fig:iou}(b), FCIS+XD outperforms the other two approaches with large performance margin. The result is not surprising due to additional use of training examples by FCIS+XD to fine-tune FCN for instance segmentation. FCIS+XD, nevertheless, is hardly to scale up to cope with dataset with unknown categories of instances. Compared to weakly supervised approach PCL*+SPN, DASR* which only leverages pre-trained model shows superior performance on INSTRE and better localization accuracy on Instance-335. DASR* is more cost-effective in terms of training and generic in detecting instances of both known and unknown categories.

\subsubsection{Scalability Test}
In this experiment, the scalability of DASR* is further studied on Instance-160 by incrementally adding in one million distracting reference images. The one million distractors are crawled from Flickr. For each image, DASR* feature is extracted with the same processing flow as before. A total of \textit{7,014,819} regions are detected by DASR, while \textit{12,486,461} regions are detected by DASR*. The scalability of DASR* is studied in comparison to several state-of-the-art approaches ranging from conventional BoVW~\cite{Sivic03}, BoVW+HE~\cite{jegou2008hamming} approaches and recent approaches, R-MAC, CroW, DeepVision and FCIS+XD. 

\begin{figure}[t]
\begin{center}
\includegraphics[width=1\linewidth]{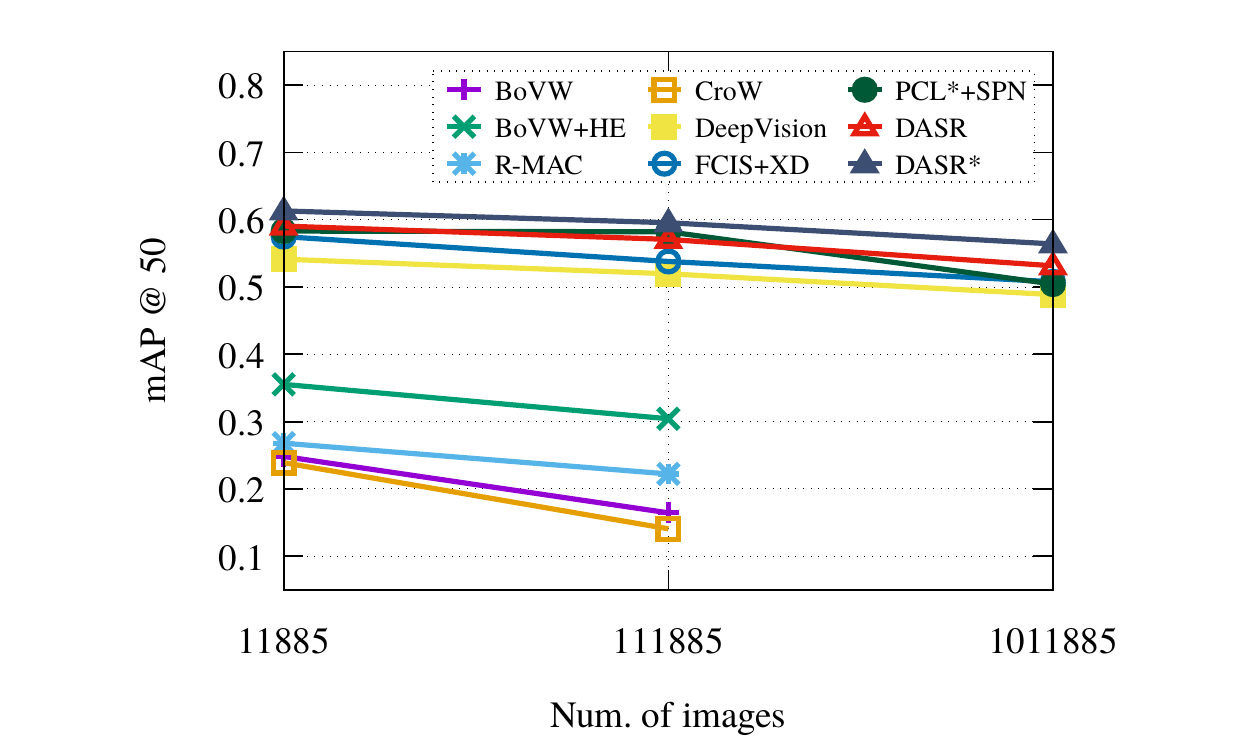}
\end{center}
   \caption{Scalability test on Instance-160 in comparison with several state-of-the-art approaches. The performance is measured by mAP@top-50. The performance is reported as a function about the number of reference images.}
\label{fig:scalability}
\end{figure}

The result is shown in Figure~\ref{fig:scalability}. Note that the performance of BoVW, BoVW+HE, R-MAC, and CroW are not reported for size beyond 100K. This is simply because their performance is already far behind DASR at the size of 100K. Overall, DASR* outperforms all the approaches, including FCIS+XD based on fully supervised model, on three different testing scales. The performance gap is mainly due to the ability of DASR* in detection the salient regions of an instance, despite that the  regions may not fully occupy the entire instance as FCIS+XD. The salient regions play an important role in guaranteeing that the features generated by DASR are more discriminative.

\subsection{Image Search Performance}
In this experiment, the effectiveness of DASR* is studied when the bag of DASR* features from one image are converted into a global image descriptor with VLAD. In our experiment, the codebook size is fixed to \textit{4}. Unlike SIFT, DASR* only needs a tiny vocabulary size since there are only \textit{12} features on average to be encoded in one image. The vocabulary is trained on an independent dataset. Following post-processing schemes proposed in~\cite{mm13:vlad}, the resulting VLAD vectors are PCA-rotated and undergone pair-wise power-law normalization with factor \textit{0.5}.

The image search performance is studied in comparison to conventional SIFT+VLAD and various of recent approaches based on deep feature. The experiments are conducted on datasets Holidays, Oxford5k, and Paris6k.  The performances of CAM, BLCF and BLCF+SalGAN on Holidays are produced based on the codes from the authors. The performance of these approaches on other two datasets are cited from the original papers. The search performance on three datasets of all the considered approaches are shown in Tab.~\ref{tab:img}.

As shown in the table, both DASR and DASR* outperform the conventional approach SIFT+VLAD considerably. In general, converting instance-level features into a global feature for retrieval will lead to performance drop. In contrast to the approaches such as R-MAC and BLCF that encode all the regional features, DASR* only rely on salient regions of instances for global image search. Therefore, the performance is expected not to be as good as the approaches based on regional feature weighting. Despite this, DASR*+VLAD is still able to achieve comparable performance as some instance-based matching approaches such as BLCF. In particular, DASR*+VLAD outperforms most of the approaches on Holidays dataset, where no dominant objects (such as landmarks) are inside the images. The satisfactory performance of DASR* on both search tasks indicate that it is possible to integrate instance and image search into one search platform, which allows the user to launch a query either about a specific instance in the image or about the image as a whole.


\begin{table}
\scriptsize
\begin{center}
\caption{Performance comparisons on three image retrieval datasets}
\label{tab:img}
\begin{tabular}{|l|r|c|c|c|}
\hline
Method & Dim. & Holidays & Oxford5k & Paris6k \\
\hline\hline
BoVW+HE~\cite{tmm16:zhao}$^{\ddag}$ & 65,536 & 0.742 & 0.503 & 0.501 \\
SIFT+VLAD*~\cite{tmm16:zhao}$^{\ddag}$ & 8,192 & 0.664 & 0.359 & 0.391 \\
R-MAC~\cite{tolias2015particular}$^{\ddag}$ & 512 & - & 0.669 & 0.830 \\
CroW~\cite{kalantidis2016cross}$^{\ddag}$ & 512 & 0.851 & 0.708 & 0.797 \\
CAM~\cite{Jimenez_2017_BMVC} & 512 & 0.785 & 0.712$^{\ddag}$ & 0.805$^{\ddag}$ \\
BLCF~\cite{mohedano2018saliency} & 336 & 0.854 & 0.722$^{\ddag}$ & 0.798$^{\ddag}$ \\
BLCF-SalGAN~\cite{mohedano2018saliency} & 336 & 0.835 & 0.746$^{\ddag}$ & 0.812$^{\ddag}$ \\
Regional Attention~\cite{kim2018regional}$^{\ddag}$ & 2,048 & - & \textbf{0.768} & \textbf{0.875} \\
DeepVision~\cite{salvador2016faster}$^{\ddag}$ & 512 & - & 0.710 & 0.798 \\ \hline
DASR+VLAD & 8,192 & 0.834 & 0.594 & 0.690 \\
DASR*+VLAD & 8,192 & \textbf{0.873} & 0.613 & 0.744 \\
\hline
\end{tabular}\\
$^{\ddag}$ digits are cited from the referred paper.

\end{center}
\end{table}

\section{Conclusion}
\label{sec:conc}
We have presented our solution for visual instance search. The focus is on the instance-level feature representation. A novel feature descriptor, namely DASR is proposed. The features are extracted from the semantically salient regions of an image that are activated by a back-propagation process. Both the instance localization and the instance-level feature description are achieved on a pre-trained classification network, without any further fine-tuning. This approach is generic in the sense that the back-propagation could be built upon any pre-trained CNN classification network. Since no fine-tune training is required, the descriptor remains effective for the instances from both known and unknown categories, which is hardly achievable with the existing approaches.

In addition to instance and image search, our approach is also potentially useful for search-driven annotation. Specifically, an annotator only needs to label a few examples of instances for an object category as queries. By our approach, all the instances of that category can be automatically retrieved from an image or video collection, with ellipses or bounding boxes indicating the instance positions. In such a way, labelling effort can be significantly reduced by requiring only to adjust the bounding boxes of instances.

\ifCLASSOPTIONcompsoc
  \section*{Acknowledgments}
\else
  \section*{Acknowledgment}
\fi

This work is supported by National Natural Science Foundation of China under grants 61572408 and 61972326, and the grants of Xiamen University 20720180074.

\bibliographystyle{IEEEtran}
\bibliography{instance}

\end{document}